\documentclass{article}
\usepackage{spconf,amsmath,graphicx,hyperref}
\usepackage{booktabs}
\usepackage{multirow}
\usepackage{xcolor}
\usepackage{subcaption}

\title{Enhancing Multi-Corpus Training in SSL-Based Anti-Spoofing Models: Domain-Invariant Feature Extraction }
\twoauthors
 {Anh-Tuan DAO, Driss Matrouf, Mickael Rouvier}
	{Laboratoire d’informatique d’Avignon, France}
 {Nicholas Evans}
	{EURECOM, Sophia Antipolis, France}

\address{}

\begin{document}
%
\maketitle
\begin{abstract}
The performance of speech spoofing detection often varies across different training and evaluation corpora. 
Leveraging multiple corpora typically enhances robustness and performance in fields like speaker recognition and speech recognition. However, our spoofing detection experiments show that multi-corpus training does not consistently improve performance and may even degrade it. We hypothesize that dataset-specific biases impair generalization, leading to performance instability.
To address this, we propose an Invariant Domain Feature Extraction (IDFE) framework, employing multi-task learning and a gradient reversal layer to minimize corpus-specific information in learned embeddings. The IDFE framework reduces the average equal error rate by 20\% compared to the baseline, assessed across four varied datasets.

\end{abstract}
\section{Introduction}
\label{sec:intro}

Automatic Speaker Verification (ASV) offers reliable biometric recognition based on voice characteristics. However, advanced spoofing attacks that mimic bona fide users pose serious security risks to ASV systems. To counter these threats, spoofing detection systems, also commonly referred to as countermeasures (CMs), are increasingly deployed to protect ASV reliability. 
In the biometrics literature, this task is formally described as Presentation Attack Detection (PAD), where the goal is to discriminate between genuine presentations and attack presentations.\footnote{In this paper, the term \emph{spoofing detection} is used interchangeably with \emph{presentation attack detection}. Similarly, \emph{bona fide} audio corresponds to \emph{genuine presentations}, while \emph{spoofing audio} refer to \emph{attack presentations}.} 
The ASVspoof~\cite{ASVspoof24} and ADD~\cite{ADD2022} communities, with broad global participation, have made substantial advances in this field.


Recent research has focused on leveraging self-supervised learning (SSL) models to enhance spoofing detection. Fine-tuning SSL representations has become the dominant paradigm, as their high-dimensional acoustic features provide superior generalization over traditional supervised end-to-end architectures \cite{AASIST,RawNet2,dao24_asvspoof}. For instance, Tak et al.~\cite{wave2vec2-seft-learning} fine-tuned the wav2vec 2.0 model, a large-scale SSL-based speech representation framework, using the ASVspoof 2019 LA dataset. Their approach demonstrated improved robustness against unseen spoofing attacks when evaluated using the ASVspoof 2021 LA and DF datasets. Roselló et al.~\cite{conformer} utilized Conformer architectures to capture both local and global speech patterns, achieving state-of-the-art performance using the same datasets. The Multi-Head Factorized Attentive (MHFA) pooling mechanism, originally developed for SSL-based speaker verification~\cite{MHFA_ASV}, has been adapted into a lightweight architecture for aggregating transformer embeddings achieving superior performance in the ASVspoof 5 Challenge~\cite{rohdin24_but_asvspoof,MHFA_Spoof}.

Despite these advancements, spoofing detection systems exhibit inconsistent performance due to dataset-specific biases in training data, as noted by Nicolas et al.~\cite{Silence2021}, who highlighted the impact of non-speech segments on model predictions.  To address these challenges, we conducted experiments to improve model reliability in the future of dataset variability. We utilized the ASVspoof 2019~\cite{ASVspoof19}, ASVspoof 5~\cite{ASVspoof24}, and Fake-or-Real (FoR)~\cite{FoR2019} datasets to create diverse training scenarios. 

In fields such as speaker recognition and speech recognition, using multiple corpora is generally known to enhance robustness and improve performance.
In contrast, our experiments in spoofing detection show that the use of multiple corpora does not consistently improve performance; in some cases, it significantly degrades performance.
Analysis of the embedding space learned under multi-corpus conditions reveals distinct clustering patterns aligned with dataset IDs, indicating the capture of corpus-specific information, potentially leading to overfitting and reduced generalization on unseen data due to dataset-specific biases. We hypothesize that these biases hinder generalization and contribute to performance instability. 

Mitigating these biases could improve generalization, enabling models to leverage multiple datasets. 
While the Gradient Reversal Layer (GRL) is a proven mechanism for mitigating domain shifts in general applications \cite{pmlr-2015-GRL}, it has also demonstrated efficacy in spoofing detection for suppressing non-discriminative attributes, such as speaker-specific identities \cite{dao2026assessingimpactspeakeridentity}.
We propose the \emph{Invariant Domain Feature Extraction} (IDFE) framework, which incorporates domain-adversarial training with a GRL to suppress dataset-specific cues in the embedding space. This approach enhances the focus on discriminative spoofing-related features leading to consistent improvement in detection performance.

Our main contributions are:
\begin{itemize}
    \item \textbf{Multi-Corpus Training and Dataset Bias Analysis}: We use four training scenarios including multi-corpus training datasets and four evaluation datasets to assess reliability. Our findings highlight significant performance variability due to dataset bias, revealing limitations in the generalization of current spoofing detection systems. 
    \item \textbf{Invariant Domain Feature Extraction (IDFE)}: We propose the IDFE framework that employs domain-adversarial training to remove corpus information from embeddings, improving generalization to unseen data distributions and reducing the impact of dataset bias on model performance. The IDFE framework achieves a 20\% reduction in average EER over four evaluation datasets compared to the MHFA baseline and can be applied to other detection architectures.

\end{itemize}

\section{SSL-based Spoofing Detection}



Our spoofing detection approach uses SSL models to extract robust audio features. It includes two components: SSL-based feature extractors and back-end classifiers.

\subsection{SSL Feature Extractors}

SSL models learn meaningful representations from unlabeled data through pretext tasks, capturing intrinsic patterns. In speech processing, SSL models like wav2vec 2.0 excel at extracting high-quality features from raw audio. The model consists of a convolutional neural network (CNN) feature encoder and a Transformer-based context network. The feature encoder processes raw audio into latent representations $z_{1:T}$, reducing dimensionality and capturing acoustic features. These representations are then input into a Transformer-based context network, which models temporal dependencies and contextual information, producing context representations $o_{1:T}$. The Transformer's self-attention mechanism enables the model to capture long-range relationships, essential for understanding speech structure.

\subsection{Back-end Classifiers}

We use three back-end classifiers as baselines:
\begin{itemize}
    \item \textbf{AASIST}~\cite{wave2vec2-seft-learning}: is specifically designed for spoofing detection, integrating spectro-temporal feature extraction with deep learning. It leverages convolutional and attention-based mechanisms to capture discriminative patterns in audio signals.
    \item \textbf{Conformer}~\cite{conformer}: combines convolutional and Transformer layers, effectively modeling both local and global dependencies in speech signals. This hybrid structure has shown strong performance in spoofing detection.
    \item \textbf{MHFA}~\cite{MHFA_Spoof}: Multi-Head Factorized Attentive (MHFA) pooling is an attention-based backend designed to efficiently aggregate Transformer embeddings using multi-head attention mechanisms with factorized attentive pooling. Originally developed for SSL-based speaker verification, MHFA employs a lightweight yet effective architecture to enhance feature representation.
\end{itemize}





\begin{figure}[t]
\centering
\includegraphics[width=1.0\linewidth]{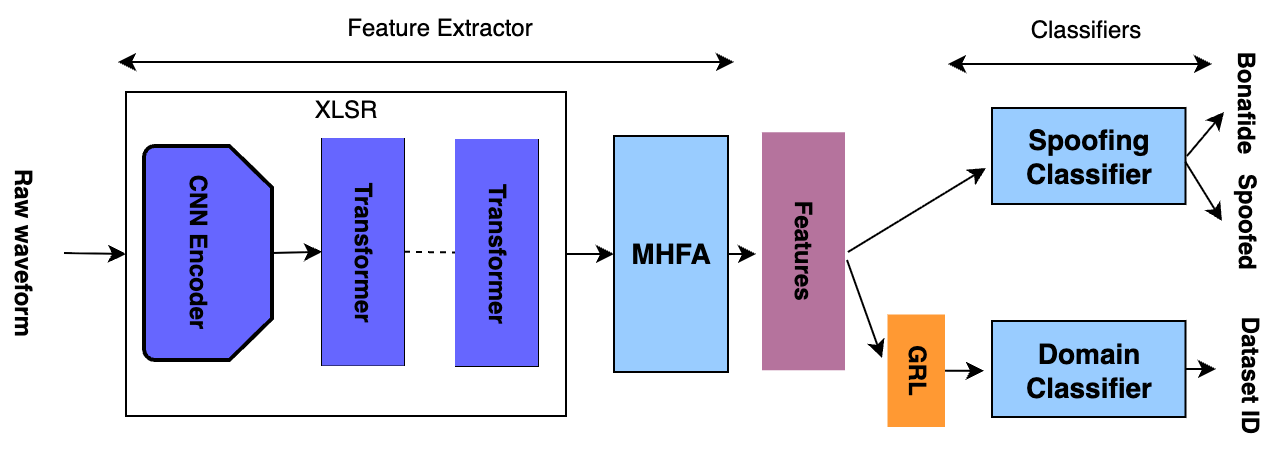}
\caption{Our model architecture with dual classifiers and GRL.}
\label{fig:idfe_arch}
\end{figure}

\section{Invariant Domain Feature Extraction}
To mitigate the negative impact of training corpus-specific artifacts, we introduce the IDFE framework. Our approach encourages the model to discard dataset-dependent information from its learned embeddings, encouraging it to focus on attack-related, discriminative cues.
\subsection{Architecture}
Our IDFE framework builds upon the SSL-based backbone and MHFA module. As illustrated in Figure~\ref{fig:idfe_arch}, the model consists of the following components.

\begin{itemize}
\item \textbf{Feature Extractor:} A pre-trained XLSR~\cite{XLS-R2022} SSL encoder to extract contextualized frame-level embeddings from raw audio, combined with a MHFA module to aggregate contextualized embeddings using attention mechanisms.
\item \textbf{Spoofing Classifier Head:} A fully connected network comprising linear layers with batch normalization, ReLU activations, and dropout, which use MHFA-aggregated embeddings to predict the binary label $\hat{y}_s \in \{\text{bona fide}, \text{spoof}\}$.
\item \textbf{Domain Classifier Head:} A parallel fully connected network (structurally similar to the spoofing head), which predicts the dataset ID $\hat{y}_d \in {1, \dots, D}$, where $D$ is the number of training corpora.
\item \textbf{Gradient Reversal Layer (GRL):} Inserted between the feature extractor and the domain classifier head. In the forward pass, the GRL acts as an identity. During backpropagation, it multiplies the gradients by a hyper parameter $-\lambda$, thus encouraging the feature extractor to remove domain-specific patterns.
\end{itemize}


\subsection{Domain Adversarial Training}

To induce domain‐invariant embeddings, we adopt a domain adversarial objective. Let $o = f(x)$ denote the feature sequence for the input audio $x$, and let $g_s(\cdot)$ and $g_d(\cdot)$ be the spoofing and domain classifiers, respectively. During training, we optimize:

\begin{equation}
\min_{f,\,g_s\, g_d}
\mathcal{L}_s\bigl(g_s(f(x)),\,y_s\bigr)
\;+\;\alpha\,
\mathcal{L}_d\bigl(g_d\bigl(\mathrm{GRL}(f(x))\bigr),\,y_d\bigr)
\label{eq1}
\end{equation}

\noindent where, $\mathcal{L}_s$ is the cross‐entropy loss for spoofing classification. $\mathcal{L}_d$ is the cross‐entropy loss for domain classification (dataset ID). The hyper parameter $\alpha$ is used to balance the two objectives, controlling the strength of domain invariance.

\label{sec:sinmt}

\begin{table*}[h]
\renewcommand{\arraystretch}{1.0} 
\centering
\caption{Model performance of baselines across four training cases. The \textit{pooled} indicates the average EER across all evaluation datasets. Details of the datasets used in each training case are provided in Section~\ref{section:4.1}.}
\begin{tabular}{llcccccc}
\toprule
& & \multicolumn{5}{c}{EER} \\
\midrule
\textbf{Training case} & \textbf{Model} & \textbf{ITW} & \textbf{ASV5 eval} & \textbf{ASV21LA-Hidden} & \textbf{ASV21DF-Hidden} & \textbf{Pooled} \\

\midrule
\multirow{3}{*}{Case 1} & {AASIST} & 7.03 & 5.54 & 13.66 & 9.6 & 8.95 \\
 & {Conformer} & 5.69	&3.85	&12.49	&10.4 & 8.10 \\
 & {MHFA} & 4.31 & 4.64 & 12.14 & 8.58 & 7.41 \\

\midrule
\multirow{3}{*}{Case 2} & AASIST & 4.05 & 12.35 & 8.9 & 8.35 & 8.41 \\
 & Conformer & 3.36	&15.12	&7.17	&7.27 & 8.23 \\
 & MHFA & 2.67 & 12.68 & 9.33 & 7.27 & 7.98 \\                               
\midrule
\multirow{3}{*}{Case 3} 
                       & MHFA & 2.27 & 8.72 & 7.57 & 5.91 & 6.11 \\
                       & MHFA-IDFE & \textbf{2.04} & \textbf{7.97} & \textbf{5.22} & \textbf{4.29} & \textbf{4.88}  \\
\midrule
\multirow{3}{*}{Case 4} 
                       & MHFA & 2.33 & 8.36 & 6.80 & 4.44 & 5.48 \\
                       & MHFA-IDFE & \textbf{1.86} & \textbf{7.95} & \textbf{5.40} & \textbf{3.58} & \textbf{4.69} \\
\bottomrule

\end{tabular}
\label{table1}
\end{table*}

\begin{table*}[h]
\renewcommand{\arraystretch}{0.9} 
\centering
\caption{Performance for MHFA-IDFE with different values of $\alpha$ which controls the
strength of domain invariance.}
\begin{tabular}{lllcccccc}
\toprule
\textbf{Training case} & \textbf{Model} & $\alpha$ & \textbf{ITW} & \textbf{ASV5 eval} & \textbf{ASV21LA-Hidden} & \textbf{ASV21DF-Hidden} & \textbf{Pooled} \\
\midrule
\multirow{3}{*}{Case 4} 
                       
                       & MHFA-IDFE & 0.1 & 1.86 & 7.95 & 5.40 & 3.58 & 4.69 \\
                       & MHFA-IDFE & 0.5 & 2.31 & 9.10 & 6.28 & 4.04 & 5.43 \\
\bottomrule
\end{tabular}
\label{table3}
\end{table*}

\subsection{Backpropagation and Parameter Updates}

During training, the model parameters are updated using stochastic gradient descent. 
Let $\theta_f$, $\theta_{g_s}$, and $\theta_{g_d}$ represent the parameters of the feature extractor, spoofing classifier, and domain classifier, respectively. 
The parameter updates are performed as follows:

\begin{equation}
\theta_f \leftarrow \theta_f - \mu \left( \frac{\partial \mathcal{L}_s}{\partial \theta_f} - \alpha \lambda \frac{\partial \mathcal{L}_d}{\partial \theta_f} \right)
\label{eq2}
\end{equation}

\begin{equation}
\theta_{g_s} \leftarrow \theta_{g_s} - \mu \frac{\partial \mathcal{L}_s}{\partial \theta_{g_s}}
\label{eq3}
\end{equation}

\begin{equation}
\theta_{g_d} \leftarrow \theta_{g_d} - \mu \frac{\partial \mathcal{L}_d}{\partial \theta_{g_d}}
\label{eq4}
\end{equation}

\noindent In Equations (\ref{eq2}-\ref{eq4}), $\mu$ denotes the learning rate, whereas $\lambda$ is a trade-off hyperparameter in GRL which controls the influence of the domain loss~\cite{pmlr-2015-GRL}. 
Equation~\ref{eq2} updates the feature extractor to reduce the spoofing loss $\mathcal{L}_s^i$ while simultaneously increasing the domain loss $\mathcal{L}_d^i$, thereby confusing the domain classifier while encouraging the spoofing classifier to learn spoofing-related cues. 
This is achieved through the GRL, which multiplies the gradient from the domain classifier by $-\lambda$ during backpropagation.
Equations~\ref{eq3} and~\ref{eq4} update the spoofing and domain classifiers, respectively, using standard gradient descent to minimize $g_s(\cdot)$ and $g_d(\cdot)$. 




\section{Experimental Setup}
\subsection{Datasets and Metrics}\label{sec:datasets-training}
\label{section:4.1}
For training, we use the ASVspoof 5 training set, the ASVspoof 2019~\cite{ASVspoof19} and FoR~\cite{FoR2019} datasets. To support multi-corpus evaluation, the ASVspoof 5 validation set is replaced with the ITW dataset. We merge the ASVspoof 2019 training and development sets to increase data volume, ensuring evaluation integrity, our evaluation sets remain disjoint and unaffected by development data leakage. We define four distinct training scenarios: 

\begin{itemize}
    \item Case 1: ASVspoof 5 training set only.
    \item Case 2: ASVspoof 2019 train-dev set.
    \item Case 3: Combination of ASVspoof 5 and ASVspoof 2019.
    \item Case 4: Combination of ASVspoof 5, ASVspoof 2019, FoR.
\end{itemize}

For evaluation, we use three datasets: ASVspoof 2021 LA and DF hidden subsets~\cite{ASVspoof21hid, ASVspoof21} and ASVspoof 5 evaluation set. The ASVspoof 2021 LA and DF hidden subsets have non-speech segments removed, preventing models from exploiting such shortcuts. Prior work~\cite{Martin-DonasARG24} shows the challenges presented by these hidden subsets, compared to standard evaluation subsets, since they demand a reliance on core spoofing cues rather than non-speech-related shortcuts. 

Performance is evaluated using the equal error rate (EER). In biometric evaluation practice, system performance is often reported in terms of the attack and bona fide presentation classification error rates (APCER and BPCER). In this paper, the EER corresponds to the operating point at which these two error rates are equal.

\subsection{Data Augmentation}
We applied standard data augmentation techniques during training, utilizing the MUSAN corpus and the real room impulse response (RIR) database~\cite{MUSAN,Reverb2017}. Each training utterance underwent one of four augmentation methods: 

\begin{itemize}
    \item Reverberation: Utterances are convolved with real RIRs to simulate reverberation effects associated with propagation in various acoustic spaces.
    \item Speech: A summation of three to eight different-speaker utterances are added to each training utterance at signal-to-noise ratios (SNRs) of 13-20~dB.
    \item Music: Randomly-selected music recordings from MUSAN are added to each training utterance at SNRs of 5-15~dB.
    \item Noise: Randomly-selected noise recordings from MUSAN are added to each training utterance at SNRs of 0-15~dB.
\end{itemize}

\begin{figure*}[t]
    \centering
    \begin{subfigure}[b]{0.33\linewidth}
        \centering
        \includegraphics[width=0.99\linewidth]{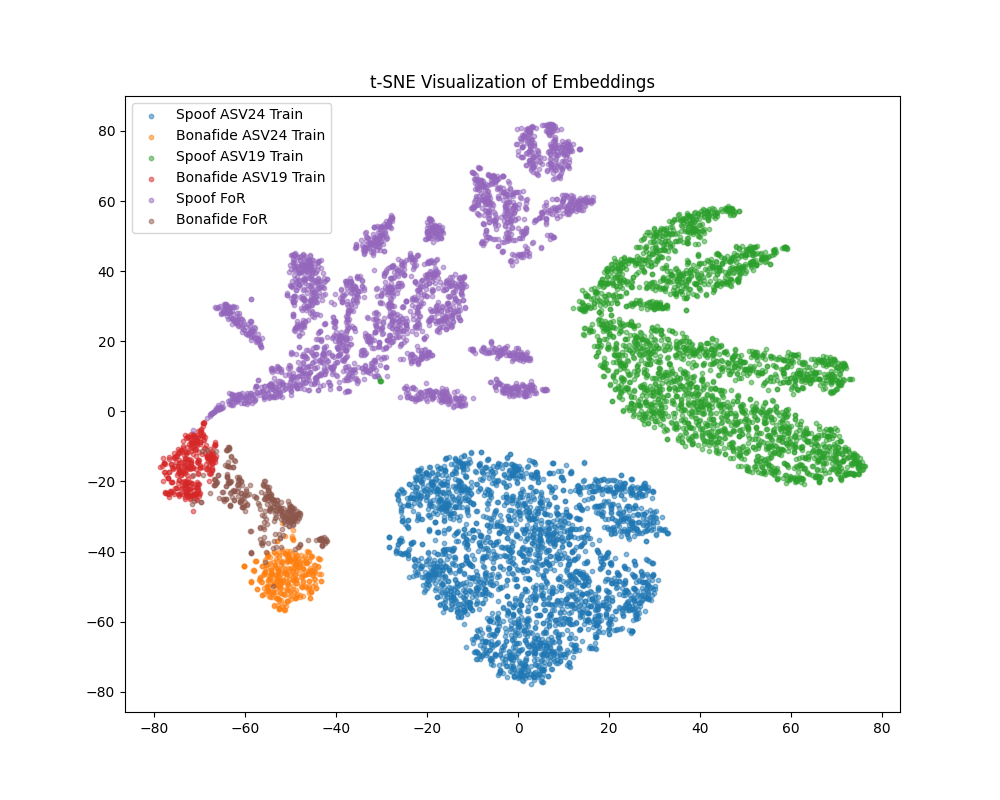}
        \caption{MHFA.}
        \label{fig:learned_emb_a}
    \end{subfigure}
    \hfill
    \begin{subfigure}[b]{0.33\linewidth}
        \centering
        \includegraphics[width=0.99\linewidth]{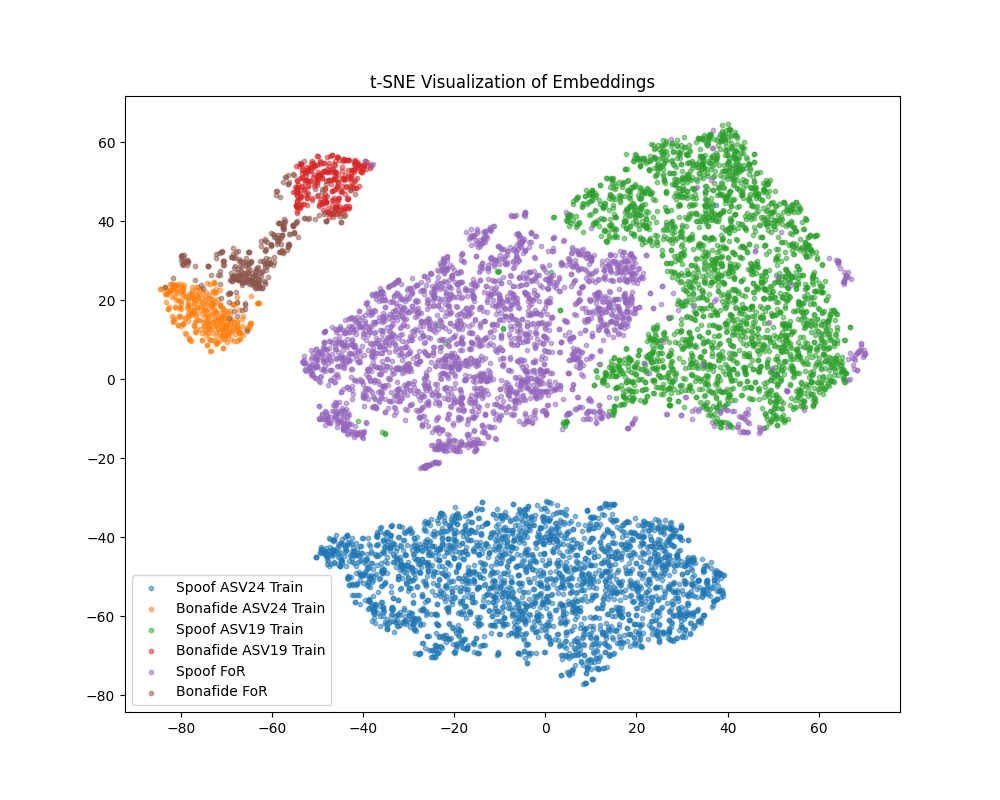}
        \caption{MHFA-IDFE ($\alpha$ = 0.1).}
        \label{fig:learned_emb_b}
    \end{subfigure}
    \hfill
    \begin{subfigure}[b]{0.33\linewidth}
        \centering
        \includegraphics[width=0.99\linewidth]{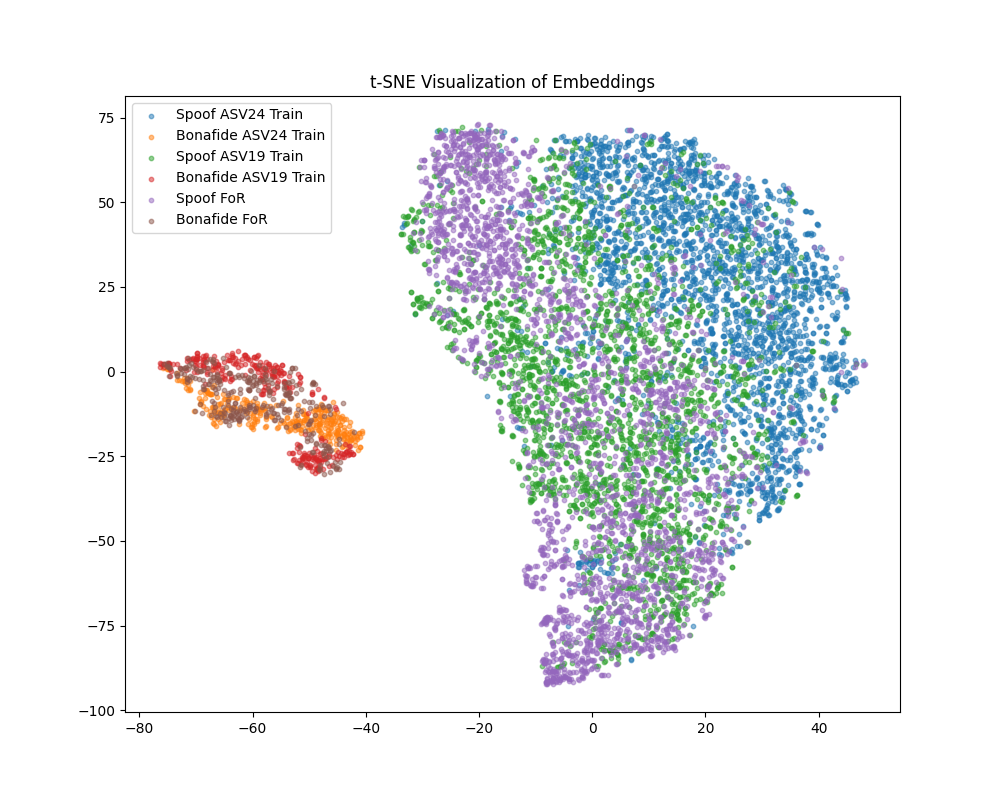}
        \caption{MHFA-IDFE ($\alpha$ = 0.5).}
        \label{fig:learned_emb_c}
    \end{subfigure}
    \caption{t-SNE visualization of learned embeddings from the MHFA and MHFA-IDFE models for the training case 4. The parameter $\alpha$ controls the strength of domain invariance, as defined in Equation~\ref{eq1}.}
    \label{fig:learned_emb}
\end{figure*}
\subsection{Implementation Details}\label{sec:implem_details}
To mitigate the non-speech bias in model training, we trim non-speech segments from the beginning and end of training utterances using the \texttt{librosa} library. Specifically, we apply \texttt{librosa.effects.trim} with a reference threshold of 40 dB to determine the duration of leading and trailing non-speech segments. Training utterances are randomly segmented into 4-second clips. For evaluation, models operate on full utterances. We use the Adam optimizer~\cite{adam} with a learning rate of $10^{-6}$ to minimize the weighted cross-entropy loss, adjusted based on the ratio of bona fide to spoofed audio samples. The hyperparameter $\alpha$ is set to 0.1 to control the strength of domain invariance. The hyperparameter $\lambda$ gradually changes from 0 to 1, following the same schedule as in \cite{pmlr-2015-GRL}. Training is conducted in batches of 32 samples over 30 epochs using NVIDIA A100 GPUs.

\section{Results}

We present an evaluation of three baseline models across two training scenarios, highlighting variability due to training and evaluation datasets. Subsequently, we select the best-performing baseline, MHFA, to integrate with our proposed IDFE framework for multi-corpus training. 

\subsection{Dataset Bias Analysis}
A performance comparison for models trained using ASVspoof 5 and ASVspoof 2019 presented in Table~\ref{table1} (Case 1-2) shows significant variations that highlight dataset bias issues. 
For instance, the MHFA model trained using ASVspoof 5 (Case 1) achieves an EER of 4.64\% for the ASVspoof 5 evaluation dataset while performance degrades to 12.14\% for the ASVspoof 2021 LA dataset. Training using ASVspoof 2019 train-dev (Case 2) yields an EER of 9.33\% for the ASVspoof 2021 LA dataset and 12.68\% for the ASVspoof 5 dataset. Similar trends are observed for AASIST and Conformer models, underscoring vulnerability to out-of-domain data. For multi-corpus training (Case 3), the combination of ASVspoof 5 and ASVspoof 2019, the MHFA model shows improved generalization for all datasets (e.g., ITW, ASVspoof 2021 LA, and DF). However, performance for the ASVspoof 5 dataset drops from 4.64\% to 8.72\% EER, indicating sensitivity to dataset composition. Among the baselines, MHFA demonstrates superior performance, making it the candidate for applying our IDFE framework in multi-corpus training.

\subsection{Multi-Corpus Training Performance}
We evaluate the MHFA-IDFE model, incorporating our IDFE framework into the MHFA baseline in two multi-corpus training scenarios (Case 3-4 in Table~\ref{table1}). In Case 3 (ASVspoof 5 + ASVspoof 2019), MHFA-IDFE performance surpasses that of MHFA across all evaluation datasets, reducing the pooled EER by 20\% (from 6.11\% to 4.88\%). The largest improvement is for the ASVspoof 2021 LA, with a 31\% EER reduction (from 7.57\% to 5.22\%). In Case 4 (ASVspoof 5, ASVspoof 2019, and FoR), MHFA-IDFE achieves a 14.4\% performance gain, reducing the pooled EER from 5.48\% to 4.69\%. Notable improvements occur for ITW and ASVspoof 2021 LA datasets, with EER reductions of 20.1\% and 20.5\%, respectively. Although MHFA-IDFE improves performance for ASVspoof 5 in Case 4 (EER from 8.36\% to 7.95\%), the EER remains higher than that in Case 1. This indicate that while MHFA-IDFE mitigates the dataset bias issue, it does not fully resolve it.

\subsection{Visualization}
A visualisation of MHFA and MHFA-IDFE embeddings for Case 4 derived using t-SNE is inlustrated in (Fig~\ref{fig:learned_emb}). 
For the MHFA model, there is a distinct clustering by dataset, suggesting the model captures corpus-specific cues. With $\alpha=0.1$, MHFA-IDFE yields embeddings with compact dataset-based clusters and clear separation between bona fide and spoofed samples; spoofed samples from FoR and ASVspoof 2019 are closer but still separable. The MHFA-IDFE with $\alpha=0.5$ produces embeddings with no evident dataset clustering, though performance is inferior to $\alpha=0.1$ (Table \ref{table3}), possibly due to excessive influence from the invariant domain loss. This excessive influence can suppress task-relevant information, such as attack type, reducing the model ability to distinguish bona fide from spoofed samples. Balancing the strength of invariant domain loss is therefore crucial to retain discriminative features while mitigating dataset bias.

\section{Conclusion}
By investigating spoofing detection in multi-corpus training, this study shows dataset bias issues, notably in the ASVspoof 5 evaluation set, emphasizing the necessity for bias mitigation. Our proposed IDFE framework fosters domain-invariant embeddings, achieving a 20\% reduction in pooled EER across four evaluation datasets. The framework effectively removes corpus-specific information from the learned embedding space. Balancing the impact of the invariant domain loss is essential to prevent excessive information loss, which could impair spoofing detection performance. 
Notably, this work emphasizes domain robustness rather than optimizing peak performance on a single backend architecture; consequently, direct comparisons with results of prior works, often trained and evaluated under different protocols, are infeasible. 


\section{Acknowledgements}

This work was performed using HPC resources from GENCI-IDRIS. This work was financially supported by ANR BRUEL (ANR-22-CE39-0009).

\bibliographystyle{IEEEbib}
\bibliography{strings,refs}

\end{document}